\documentclass[runningheads]{llncs}
\usepackage{graphicx}
\usepackage{graphicx}  
\usepackage{amsmath}
\usepackage{amssymb}
\usepackage[linesnumbered,ruled]{algorithm2e}
\usepackage{algpseudocode}
\usepackage{subfig}

\newcommand{\act}[1]{#1}

\newcommand{\pref}{\succ}
\newcommand{\state}[1]{#1}
\newcommand{\matr}[1]{\mathbf{#1}}

\newcommand{\citep}[1]{}
\newcommand{\citet}[1]{}

\newcommand{\mathds}[1]{\mathbb{#1}}

\usepackage{pgfplots}

\usetikzlibrary{patterns}
\usepgfplotslibrary{fillbetween}
\usetikzlibrary{patterns}

\newlength{\hatchspread}
\newlength{\hatchthickness}
\newlength{\hatchshift}
\newcommand{\hatchcolor}{}
\tikzset{hatchspread/.code={\setlength{\hatchspread}{#1}},
         hatchthickness/.code={\setlength{\hatchthickness}{#1}},
         hatchshift/.code={\setlength{\hatchshift}{#1}},
         hatchcolor/.code={\renewcommand{\hatchcolor}{#1}}}
\tikzset{hatchspread=3pt,
         hatchthickness=0.4pt,
         hatchshift=0pt,
         hatchcolor=black}
\pgfdeclarepatternformonly[\hatchspread,\hatchthickness,\hatchshift,\hatchcolor]
   {custom north west lines}
   {\pgfqpoint{\dimexpr-2\hatchthickness}{\dimexpr-2\hatchthickness}}
   {\pgfqpoint{\dimexpr\hatchspread+2\hatchthickness}{\dimexpr\hatchspread+2\hatchthickness}}
   {\pgfqpoint{\dimexpr\hatchspread}{\dimexpr\hatchspread}}
   {
    \pgfsetlinewidth{\hatchthickness}
    \pgfpathmoveto{\pgfqpoint{0pt}{\dimexpr\hatchspread+\hatchshift}}
    \pgfpathlineto{\pgfqpoint{\dimexpr\hatchspread+0.15pt+\hatchshift}{-0.15pt}}
    \ifdim \hatchshift > 0pt
      \pgfpathmoveto{\pgfqpoint{0pt}{\hatchshift}}
      \pgfpathlineto{\pgfqpoint{\dimexpr0.15pt+\hatchshift}{-0.15pt}}
    \fi
    \pgfsetstrokecolor{\hatchcolor}
    \pgfusepath{stroke}
   }

\definecolor{green1}{HTML}{06634E}
\definecolor{green2}{HTML}{1B9478}
\definecolor{red1}{HTML}{692109}
\definecolor{red2}{HTML}{944023}
\definecolor{blau}{HTML}{00A3D6}
\definecolor{rot}{HTML}{D93B77}
\definecolor{occa}{HTML}{84BD00}
\definecolor{vio}{HTML}{6F3485}

\definecolor{b1}{HTML}{FF0000 }
\definecolor{b2}{HTML}{FF5500 }
\definecolor{b3}{HTML}{ff7700 }
\definecolor{b4}{HTML}{ff9900 }
\definecolor{b5}{HTML}{ffbb00 }
\definecolor{b6}{HTML}{ffdd00 }
\definecolor{b7}{HTML}{ffff00}
\definecolor{b8}{HTML}{ccff00}
\definecolor{b9}{HTML}{00ff00}

\title{Preference-Based Monte Carlo Tree Search}
\author{Tobias Joppen \and
Christian Wirth \and
Johannes F\"urnkranz}
\authorrunning{T.Joppen et al.}
\institute{Technische Universit\"at Darmstadt, Darmstadt, Germany
\email{\{tjoppen,cwirth,juffi\}@ke.tu-darmstadt.de}}

\begin{document}

\maketitle
\begin{abstract}
  Monte Carlo tree search (MCTS) is a popular choice for solving sequential anytime problems.
However, it depends on a numeric feedback signal, which can be difficult to define.
Real-time MCTS is a variant which 
may only rarely encounter states with an explicit, extrinsic reward.
To deal with such cases, the experimenter has to supply an additional numeric feedback signal in the form of a heuristic, which intrinsically guides the agent. 
%
Recent work has shown evidence that in different areas 
the underlying structure is ordinal and not numerical.
Hence erroneous and biased heuristics are inevitable, especially in such domains.
In this paper, we propose a MCTS variant which only depends on qualitative feedback, and therefore
opens up new applications for MCTS. 
We also find indications that translating absolute into ordinal feedback may be beneficial.
Using a puzzle domain, we show that our preference-based MCTS variant, wich only receives qualitative feedback, is able to reach a performance level comparable to a regular MCTS baseline, which obtains quantitative feedback.

\end{abstract}

\section{Introduction}
\label{sec:Introduction}

Many modern AI problems can be described as a Markov decision processes (MDP), where it is required to select the best action in a given state, in order to maximize the expected long-term reward.
\emph{Monte Carlo tree search (MCTS)} is a popular technique for determining the best actions in MDPs \cite{UCT,MCTSSurvey}, which combines game tree search with bandit learning.
It has been particularly successful in game playing, most notably in Computer Go \cite{RecentComputerGo}, where it was the first algorithm to compete with professional players in this domain \cite{MoGo,silver2017mastering}. 
MCTS is especially useful if no state features are available and strong time constraints exist, like in general game playing \cite{FinssonThesis} or for opponent modeling in poker \cite{AAAIW101984}.

Classic MCTS depends on a numerical feedback or reward signal, as assumed by the MDP framework, where the algorithm tries to maximize the expectation of this reward.
However, for humans it is often hard to define or to determine exact numerical feedback signals. 
Suboptimally defined reward may allow the learner to maximize its rewards without reaching the desired extrinsic goal \cite{amodei2016concrete} or may require a predefined trade-off between multiple objectives \cite{Knowles2001}.

This problem is particularly striking in settings where the natural feedback signal is inadequate to steer the learner to the desired goal. For example, if the problem is a complex navigation task and a positive reward is only given when the learner arrives in the goal state, the learner may fail because it will never find the way to the goal, and may thus never receive feedback from which it can improve its state estimations.

Real-time MCTS \cite{pepels2014real,MCTSSurvey} is a popular variant of MCTS often used in real-time scenarios, which tries to solve this problem by introducing heuristics to guide the learner. Instead of solely relying on the natural, \emph{extrinsic} feedback from the domain, it assumes an additional \emph{intrinsic} feedback signal, which is comparable to the heuristic functions commonly used in classical problem solving techniques.
In this case, the learner may observe intrinsic reward signals for non-terminal states, in addition to the extrinsic reward in the terminal states.
Ideally, this intrinsic feedback should be designed to naturally extend the extrinsic feedback, reflecting the expected extrinsic reward in a state, but this is often a hard task. In fact, if perfect intrinsic feedback is available in each state, making optimal decisions would be trivial.
%
Hence heuristics are often error-prone
and may lead to suboptimal solutions in that MCTS may get stuck in locally optimal states. 
Later we introduce heuristic MCTS (H-MCTS), which uses this idea of evaluating non-terminal states with heuristics but is not bound to real-time applications.

On the other hand, humans are often able to provide reliable qualitative feedback.
In particular,
humans tend to be less competent in providing exact feedback values on a numerical scale than to determine
the better
of
two states in a pairwise comparison \cite{Thurstone}.
%
This observation forms the basis of \emph{preference learning}, which is concerned with learning ranking models from such qualitative training information \cite{plbook}.
Recent work has presented and supported the assumption that emotions are by nature relative and similar ideas exist in topics like psychology, philosophy, neuroscience, marketing research and more \cite{yannakakis2017ordinal}.
Following this idea, extracting preferences from numeric values does not necessarily mean a loss of information (the absolute difference), but a loss of biases caused through absolute annotation \cite{yannakakis2017ordinal}.
Since many established algorithms like MCTS are not able to work with preferences, modifications of algorithms have been proposed to enable this, like in the realm of reinforcement learning \cite{jf:MLJ-PrefBasedRL,PrefEpisodicREPS,DeepPBRL}.

In this paper we propose a variant of MCTS which works on ordinal reward MDPs (OMDPs) \cite{weng2011markov}, instead of MDPs.
\begin{figure}[t]
\centering
\includegraphics[width=0.7\columnwidth]{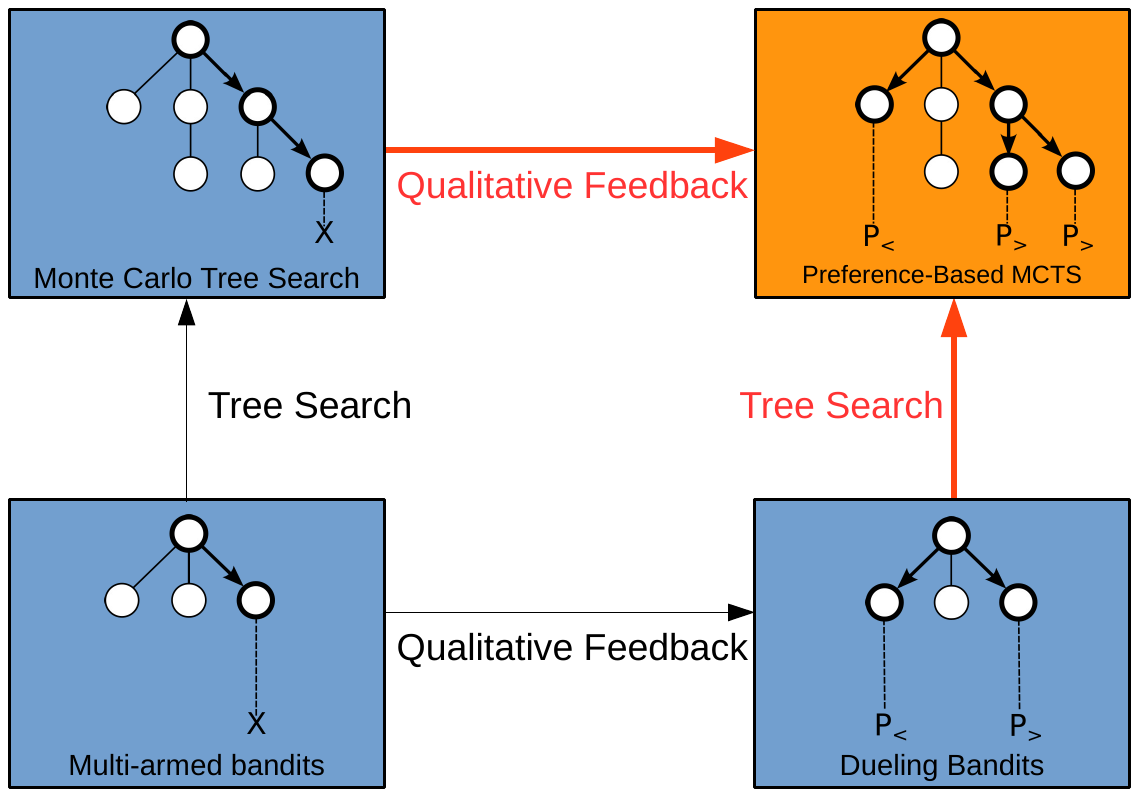}
\caption{Research in Monte Carlo methods}
\label{fig:MCTSResearch}
\end{figure}
%
%
%
%
%
The basic idea behind the resulting preference-based Monte Carlo tree search algorithm is to use the principles of \emph{preference-based} or \emph{dueling bandits} \cite{yue2009interactively,DuelBanditJournal,PrefBanditSurvey} to replace the \emph{multi-armed bandits} used in classic MCTS. Our work may thus be viewed as either extending the work on preference-based bandits to tree search, or to extend MCTS to allow for preference-based feedback, as illustrated in 
Fig.~\ref{fig:MCTSResearch}.
Thereby, the tree policy does not select a single path, but a binary tree leading to multiple rollouts per iteration and we obtain pairwise feedback for these rollouts.

We evaluate the performance of this algorithm by comparing it to heuristic MCTS (H-MCTS). 
Hence, we can determine the effects of approximate, heuristic feedback in relation to the ground truth. 
We use the 8-puzzle domain since simple but imperfect heuristics already exist for this problem.
In the next section, we start the paper with an overview of MDPs, MCTS and preference learning.


\section{Foundations}

In the following, we review the concepts of \emph{Markov decision processes} (MDP), \emph{heuristic Monte Carlo tree search} (H-MCTS) and \emph{preference-based bandits}, which form the basis of our work. We use an MDP as the formal framework for the problem definition, and H-MCTS is the baseline solution strategy we build upon. We also briefly recapitulate \emph{multi armed bandits} (MAP) as the basis of MCTS and their extension to preference-based bandits.

\subsection{Markov Decision Process}
A typical Monte Carlo tree search problem can be formalized as a Markov Decision Process (MDP) \cite{MDP}, consisting of
a set of \emph{states} $S$,
the set of \emph{actions} $A$ that the agent can perform (where $A(\state{s}) \subset A$ is applicable in state $s$),
a \emph{state transition} function $\delta(\state{s}'\mid \state{s},\act{a})$,
a \emph{reward function} $r(\state{s}) \in \mathbb{R}$ for reaching state $s$ and
a distribution $\mu(\state{s}) \in [0,1]$ for starting states.
We assume a single \emph{start state} and non-zero rewards only in terminal states.

An Ordinal Reward MDP (OMDP) is similar to MDP but the reward function, which does not lie in $\mathbb{R}$, but is defined over a qualitative scale,
such that states can only be compared preference wise.

The task is to learn a \emph{policy} $\pi(\act{a}\mid \state{s})$ that defines the probability of selecting an action $\act{a}$ in state $\state{s}$. The optimal policy $\pi^*(\act{a}\mid \state{s})$ maximizes the expected, cumulative reward \cite{ReinforcementLearning} (MDP setting), or maximizes the preferential information for each reward in the trajectory \cite{weng2011markov} (OMDP setting).
For finding an optimal policy, one needs to solve the so-called exploration/exploitation problem. The state/action spaces are usually too large to sample exhaustively. Hence, it is required to trade off the improvement of the current, best policy (exploitation) with an exploration of unknown parts of the state/action space.

\begin{figure}[t]
	\centering
	\includegraphics[width=1\columnwidth]{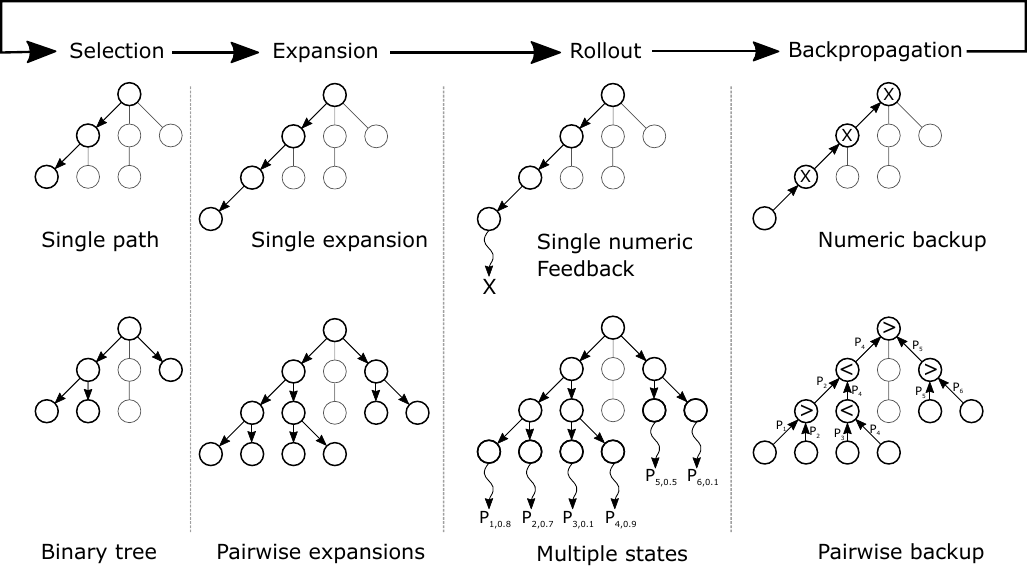}
	\caption{Comparisons of MCTS (top) and preference-based MCTS (bottom)}
	\label{fig:MCTSCompare}
\end{figure}
\subsection{Multi-armed Bandits}
\emph{Multi-armed bandits (MABs)} are a method for identifying the arm (or action) with the highest return by repeatedly pulling one of the possible arms.
They may be viewed as an MDP with only one non-terminal state, and 
the task is to achieve the highest average reward in the limit.
Here the exploration/exploitation dilemma is to play the best-known arm often (exploitation) while it is at the same time necessary to search for the best arm (exploration).
A well-known technique for resolving this dilemma in bandit problems are \emph{upper confidence bounds} (UCB \cite{UCB}), which allow to bound the expected reward for a certain arm, and to choose the action with the highest associated upper bound. The bounds are iteratively updated based on the observed outcomes. The simplest UCB policy 
\begin{equation}
UCB1 = \bar{X_j} + \sqrt{\frac{2 \ln n}{n_j}}
\label{eq:ucb}
\end{equation}
\noindent
adds a bonus of $\sqrt{2 \ln n / n_j}$, based on the number of performed trials $n$ and how often an arm was selected ($n_j$). The first term favors arms with high payoffs, while the second term guarantees exploration \cite{UCB}. 
The reward is expected to be bound by $[0,1]$.


\subsection{Monte Carlo Tree Search}

Considering not only one but multiple, sequential decisions leads to sequential decision problems. 
\emph{Monte Carlo tree search} (MCTS) is a method for approximating an optimal policy for a MDP. 
It builds a partial search tree, guided by the estimates for the encountered actions \cite{UCT}.
The tree expands deeper in parts with the most promising actions and spends less time evaluating less promising action sequences. 
The algorithm iterates over four steps, illustrated in the upper part of Fig.~\ref{fig:MCTSCompare} \cite{MCTSSurvey}:

\begin{enumerate}
\item \emph{Selection:} Starting from the initial state $\state{s}_0$, a \emph{tree policy} is applied until a state is encountered that has unvisited successor states.
\item \emph{Expansion:} One successor state is added to the tree.
\item \emph{Simulation:} Starting from this state, a \emph{simulation policy} is applied until a terminal state is observed.
\item \emph{Backpropagation:} The reward accumulated during the simulation process is backed up through the selected nodes in tree.
\end{enumerate}

In order to adapt UCB to tree search, it is necessary to consider a bias, which results from the uneven selection of the child nodes, in the tree selection policy. The UCT policy
\begin{equation}
UCT = \bar{X_j} + 2 C_p \sqrt{\frac{2 \ln n}{n_j}}
\label{eq:uct}
\end{equation}
\noindent
has been shown to be optimal within the tree search setting up to a constant
factor \cite{UCT}.

\subsection{Heuristic Monte Carlo Tree Search}
In large state/action spaces, rollouts can take many actions until a terminal state is observed. However, long rollouts are subject to high variance due to the stochastic sampling policy. Hence, it can be beneficial to disregard such long rollouts in favor of shorter rollouts with lower variance. 
\emph{Heuristic MCTS} (\mbox{H-MCTS}) stops rollouts after a fixed number of actions and uses a heuristic evaluation function in case no terminal state was observed \cite{pepels2014real,perez2016multi}. 
The heuristic is assumed to approximate $V(\state{s})$ and can therefore be used to update the expectation.



\subsection{Preference-Based Bandits}
\label{sec:PB-MAB}



\emph{Preference-based multi-armed bandits} (PB-MAB), closely related to \emph{dueling bandits}, are the adaption of multi-armed bandits to preference-based feedback \cite{yue2009interactively}. 
Here the bandit iteratively chooses two arms that get compared to each other. 
The result of this comparison is a preference signal that indicates which of two arms $\act{a}_i$ and $\act{a}_j$ is the better choice ($\act{a}_i \pref \act{a}_j$) or whether they are equivalent. 


\label{sec:RUCB}
The \emph{relative UCB} algorithm (RUCB \cite{zoghi14}) allows to compute approximate, optimal policies for PB-MABs by computing the Condorcet winner,  
i.e., the action that wins all comparisons to all other arms.
To this end, RUCB stores the number of times $w_{ij}$ an arm $i$ wins against another arm $j$ and uses this information to calculate an upper confidence bound 
\begin{equation}u_{ij} = \frac{w_{ij}}{w_{ij} + w_{ji}} + \sqrt{\dfrac{\alpha \operatorname{ln}t}{w_{ij} + w_{ji}}},
  \label{eq:RUCB}
\end{equation}
for each pair of arms. $\alpha > \frac{1}{2}$ is a parameter to trade-off exploration and exploitation and $t$ is the number of observed preferences.
These bounds are used to maintain a set of possible Condorcet winners.
If at least one possible Condorcet winner is detected, it is tested against its hardest competitor.

Several alternatives to RUCB have been investigated in the literature, but most PB-MAB algorithms are "first explore, then exploit" methods. They explore until a pre-defined number of iterations is reached, and start exploiting afterwards.  Such techniques are only applicable if it is possible to define the number of iterations in advance. 
But this is not possible to do for each node. Therefore we use RUCB in the following.
For a general overview of PB-MAB algorithms, we refer the reader to \cite{PrefBanditSurvey}.



\section{Preference-Based Monte Carlo Tree Search}
\label{sec:PB-MCTS}

\begin{figure}[t]
	\centering
	\includegraphics[width=300pt]{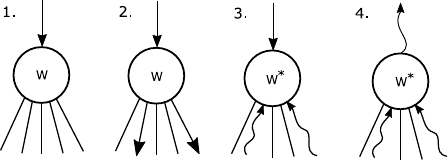}
	\caption{A local node view of PB-MCTS's iteration:
	selection; child selection; child backprop and update; backprop one trajectory.}
	\label{fig:step}
\end{figure}

\begin{algorithm}[t]
	\SetKwInOut{Input}{Input}
	\SetKwInOut{Output}{Output}
	
	\textbf{function PB-MCTS} $(T, \state{s}, \alpha, \matr{W}, B)$\;
	\Input{A set of explored states $\hat{S}$, the current state $\state{s}$,
		exploration-factor $\alpha$, matrix of wins $\matr{W}$ (per state),
		list of last Condorcet pick $B$ (per state)}
	\Output{$[\state{s}', \hat{S}, \matr{W}, B]$ 
        }
	$[\act{a}_1, \act{a}_2, B] \leftarrow \textsc{SelectActionPair}(\matr{W}_{\state{s}}, B_{\state{s}})$\;
	\For{$\act{a} \in \{\act{a}_1, \act{a}_2\}$}{
		$\state{s}' \sim \delta(\state{s}' \mid \state{s},\act{a})$\;
		\eIf{$\state{s}' \in \hat{S}$}{
			$[sim[\act{a}], \hat{S}, \matr{W}, B] \leftarrow   \textsc{\quad PB-MCTS}(\hat{S}, \state{s}', \alpha, \matr{W}, B)$; 
		}
		{
			$\hat{S} \leftarrow \hat{S} \cup \{\state{s}'\}$\;
			$sim[\act{a}] \leftarrow \textsc{Simulate}(\act{a})$\;
		}
	}
	
	$w_{\state{s} \act{a}_1 \act{a}_2} \leftarrow
	 w_{\state{s} \act{a}_1 \act{a}_2} + \mathds{1}(sim[\act{a}_2] \pref sim[\act{a}_1])$
        \hspace*{24mm}$+ \frac{1}{2}\mathds{1}(sim[\act{a}_1] \simeq sim[\act{a}_2])$\;
	$w_{\state{s} \act{a}_2 \act{a}_1} \leftarrow w_{\state{s} \act{a}_2 \act{a}_1} + \mathds{1}(sim[\act{a}_1] \pref sim[\act{a}_2])$
        \hspace*{24mm}$+ \frac{1}{2}\mathds{1}(sim[\act{a}_2] \simeq sim[\act{a}_1])$\;
	$\state{s}_{\textit{return}} \leftarrow \textsc{ReturnPolicy}(\state{s}, \act{a}_1, \act{a}_2, sim[\act{a}_1], sim[\act{a}_2])$\;
	return [$\state{s}_{\textit{return}}, T, \matr{W}, B$]\;
	
	\caption{One Iteration of PB-MCTS}
	\label{alg:heuristicSearch}
\end{algorithm}

In this section, we introduce a preference-based variant of Monte Carlo tree search \mbox{(PB-MCTS)}, as shown in Fig.~\ref{fig:MCTSResearch}.
%
This work can be viewed as an extension of previous work in two ways: (1) it adapts Monte Carlo tree search to preference-based feedback, comparable to the relation between preference-based bandits and multi-armed bandits, and (2) it generalizes preference-based bandits to sequential decision problems like MCTS generalizes multi-armed bandits.

To this end, we adapt RUCB to a tree-based setting, as shown in Algorithm~\ref{alg:heuristicSearch}.
In contrast to H-MCTS, PB-MCTS works for OMDPs and selects two actions 
 per node in the \emph{selection phase}, as shown in Fig.~\ref{fig:step}.
%
Since RUCB is used as a tree policy, each node in the tree maintains its own weight matrix  $\matr{W}$ to store the history of action comparisons in this node.
Actions are then selected based on a modified version of the RUCB formula~\eqref{eq:RUCB} 
  \begin{eqnarray}
\hat{u}_{ij} &=& \frac{w_{ij}}{w_{ij}+w_{ji}}+c\sqrt{\frac{\alpha \ln t}{w_{ij}+w_{ji}}},\\
&=&\frac{w_{ij}}{w_{ij}+w_{ji}}+\sqrt{\frac{\hat{\alpha} \ln t}{w_{ij}+w_{ji}}},\nonumber
\end{eqnarray}
\noindent
where $\alpha>\frac{1}{2}$, $c>0$ and $\hat{\alpha}=c^2\alpha>0$ are the hyperparameters that allow to trade off exploration and exploitation.
Therefore, RUCB can be used in trees with the corrected lower bound $0<\alpha$.

Based on this weight matrix, \textsc{SelectActionPair} then selects two actions using the same strategy as in RUCB:
If $C \neq \emptyset$, the first action $a_1$ is chosen among the possible Condorcet winners 
$
C = \{\act{a}_c \mid \forall j:u_{cj}\geq 0.5\}.
$
Typically, the choice among all candidates $c \in C$ is random. However, 
in case the last selected Condorcet candidate in this node is still 
in $C$, it has a $50\%$ chance to be selected again, whereas each of the other candidates can be share the remaining $50\%$ of the probability mass evenly.
The second action $a_2$ is chosen to be $a_1$'s hardest competitor, i.e., the move whose win rate against $a_1$ has the highest upper bound
$
a_2 = \arg \max_l u_{la_1}.
$
Note that, just as in RUCB, the two selected arms need not necessarily be different, i.e., it may happen that $a_1 = a_2$. This is a useful property because once the algorithm has reliably identified the best move in a node, forcing it to play a suboptimal move in order to obtain a new preference would be counter-productive.
In this case, only one rollout is created and the node will not receive a preference signal in this node. However, the number of visits to this node are updated, which may lead to a different choice in the next iteration.

The \emph{expansion and simulation phases} are essentially the same as in conventional MCTS except that multiple nodes are expanded in each iteration.
\textsc{Simulate} executes the \textit{simulation policy} until a terminal state or break condition occurs as explained below. In our experiments the simulation policy performs a random choice among all possible actions.
Since two actions per node are selected, one simulation for each action is conducted in each node.
Hence, the algorithm traverses a binary subtree of the already explored state space tree before selecting multiple nodes to expand. As a result, the number of rollouts is not constant in each iteration but increases exponential with the tree depth.
The preference-based feedback is obtained from a pairwise comparison of the performed rollouts.

In the \emph{backpropagation phase}, the obtained comparisons are propagated up towards the root of the tree. 
In each node, the $\matr{W}$ matrix is updated by comparing the simulated states of the corresponding actions $i$ and $j$ and updating the entry $w_{ij}$. Passing both rollouts to the parent in each node would result in a exponential increase of pairwise comparisons, due to the binary tree traversal.
Hence, the newest iteration could dominate all previous iterations in terms of the gained information.
This is a problem, since the feedback obtained in a single iteration may be noisy and thus yield unreliable estimates.
Monte Carlo techniques need to average multiple samples to obtain a sufficient estimate of the expectation.
Multiple updates of two actions in a node may cause further problems:
The preferences may arise from bad estimates since one action may not be as well explored as the other.
It would be unusual for RUCB to select the same two actions multiple times consecutively, since either the first action is no Condorcet candidate anymore or the second candidate, the best competitor, will change.
These problems may lead to unbalanced exploration and exploitation terms resulting in overly bad ratings for some actions.
Thus, only one of the two states is propagated back to the root node.
This way it can be assured that the number of pairwise comparisons in the nodes (and especially in the root node) remains constant ($=1$) over all iterations, ensuring numerical stability.

For this reason, we need a \emph{return policy} to determine what information is propagated upwards (compare \textsc{ReturnPolicy} in Alg.~\ref{alg:heuristicSearch}).
An obvious choice is the 
\emph{best preference policy} (BPP), which always propagates the preferred alternative upwards, as illustrated in step four of Fig.~\ref{fig:step}.
A random selection is used in case of indifferent actions.
We also considered returning the best action according to the node's updated matrix $\matr{W}$, to make a random selection based on the weights of $\matr{W}$, and to make a completely random selection. However, preliminary experiments showed a substantial advantage when using BPP.






\section{Experimental Setup}
\label{sec:experiments}

  \begin{figure}[t]
\centering
	\includegraphics[width=200pt]{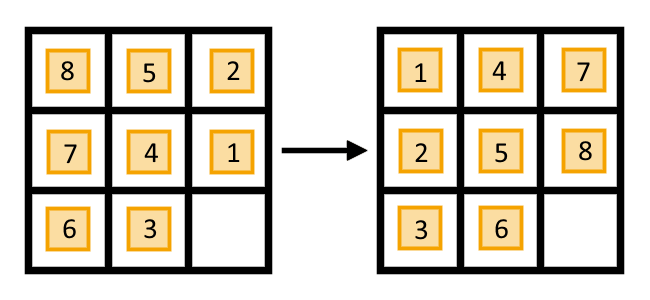}
	\caption{The start state (left) and end state (right) of the 8-Puzzle. 
		The player can swap the positions of the empty field and one adjacent number.}
	\label{fig:8puzzle}
\end{figure}

We compare PB-MCTS to H-MCTS in the 8-puzzle domain.
The 8-puzzle is a move-based deterministic puzzle where the player can move numbers on a grid.
It is played on a $3\times3$ grid where each of the $9$ squares is either blank or has a tile with number $1$ to $8$ on it.
A move consists of shifting one of the up to $4$ neighboring tiles to the blank square, thereby exchanging the position of the blank and this neighbor.
The task is then to find a sequence of moves that lead from a given start state to a known end state (see Fig.~\ref{fig:8puzzle}). 
The winning state is the only goal state. Since it is not guaranteed to find the goal state, the problem is an infinite horizon problem. However, we terminate the evaluation after $100$ time-steps to limit the runtime. Games that are terminated in this way are counted as losses for the agent. The agent is not aware of this maximum.


\subsection{Heuristics}

\begin{figure}[t]
	\subfloat[Manhattan distance]{\centering{ \includegraphics[width=170pt]{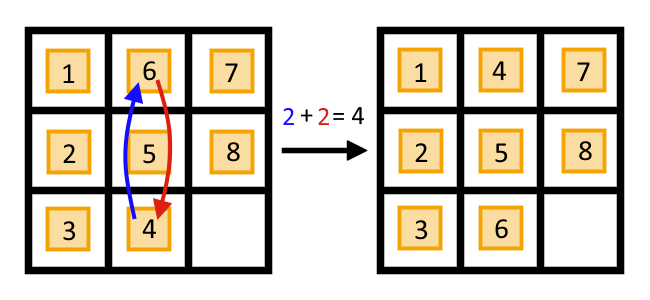}}}
	\subfloat[Manhattan distance with linear conflict\label{fig:heuristics-MDC}]{\centering{ \includegraphics[width=170pt]{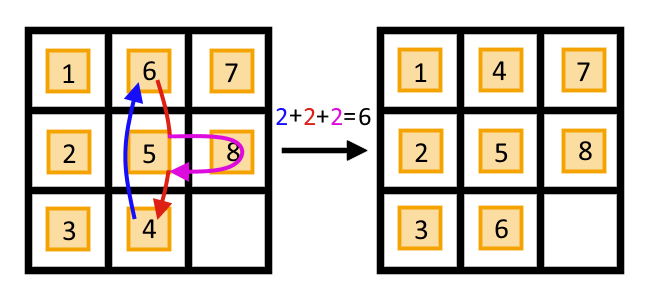}}}
	
	\caption{The two heuristics used for the 8-puzzle.}
	\label{fig:heuristics}
\end{figure}

As a heuristic for the 8-puzzle, we use the \emph{Manhattan distance with linear conflicts} (MDC), a variant of 
the \emph{Manhattan distance} (MD). 
MD is an optimistic estimate for the minimum number of moves required to reach the goal state.
It is defined as
\begin{equation}
  H_{\textit{manhattan}}(\state{s}) = \sum_{i=0}^8 |pos(\state{s},i) - goal(i)|,
\label{eq:manhattan}
\end{equation}
where $pos(\state{s},i)$ is the $(x,y)$ coordinate of number $i$ in game state $\state{s}$, $goal(i)$ is its position in the goal state, and $|\cdot|_1$ refers to the 1-norm or Manhattan-norm.


MDC additionally detects and penalizes linear conflicts.
Essentially, a linear conflict occurs if two numbers $i$ and $j$ are on the row where they belong, but on swapped positions. For example, in Fig.~\ref{fig:heuristics-MDC}, the tiles $4$ and $6$ are in the right column, but need to pass each other in order to arrive at their right squares.
For each such linear conflict, MDC increases the MD estimate by two because in order to resolve such a linear conflict,
at least one of the two numbers needs to leave its target row (1st move) to make place for the second number, 
and later needs to be moved back to this row (2nd move). The resulting heuristic is still admissible in the sense that it can never over-estimate the actually needed number of moves.

\subsection{Preferences}
In order to deal with the infinite horizons during the search, both algorithms rely on the same heuristic evaluation function, which is called after the rollouts have reached a given depth limit.
For the purpose of comparability, both algorithms use the same heuristic for evaluating non-terminal states, but PB-MCTS does not observe the exact values but only preferences that are derived from the returned values.
Comparing arm $a_i$ with $a_j$ leads to terminal or heuristic rewards $r_i$ and $r_j$, based on the according rollouts.
From those reward values, we derive preferences 
$$ (\act{a}_k \pref \act{a}_l) \Leftrightarrow (r_k > r_l) \text{ and } (\act{a}_k \simeq \act{a}_l) \Leftrightarrow (r_k = r_l)$$
which are used as feedback for PB-MCTS. H-MCTS can directly observe the reward values $r_i$.

\pgfplotscreateplotcyclelist{my black white}{%
	blau, solid, every mark/.append style={solid, fill=blau}, mark=*\\%
	rot, solid, every mark/.append style={solid, fill=rot}, mark=square*\\%
	occa, dashed, every mark/.append style={solid, fill=occa}, mark=diamond*\\%
	vio, dashed, every mark/.append style={solid, fill=vio}, mark=triangle*\\%
}

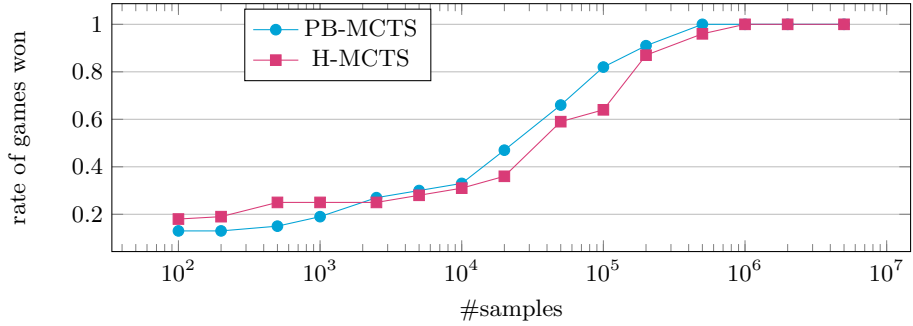
\begin{figure}[t]
	\begin{tikzpicture}
\begin{semilogxaxis}[
	height = 0.4*\linewidth,
	width = \linewidth,
	xlabel={\#samples},
	ylabel={rate of games won},
	legend style = { at = {(0.4,0.98)}},
   ymajorgrids,
	cycle list name=my black white
]

\addplot coordinates{
(100,0.130)
(200,0.130)
(500,0.150)
(1000,0.190)
(2500,0.270)
(5000,0.300)
(10000,0.330)
(20000,0.470)
(50000,0.660)
(100000,0.820)
(200000,0.910)
(500000,1.000)
(1000000,1.000)
(2000000,1.000)
(5000000,1.000)
};

\addplot coordinates{
(100,0.180)
(200,0.190)
(500,0.250)
(1000,0.250)
(2500,0.250)
(5000,0.280)
(10000,0.310)
(20000,0.360)
(50000,0.590)
(100000,0.640)
(200000,0.870)
(500000,0.960)
(1000000,1.000)
(2000000,1.000)
(5000000,1.000)
};

\legend{PB-MCTS,H-MCTS}
\end{semilogxaxis}
\end{tikzpicture}\label{fig:maxConfig}
	\caption{Using their best hyperparameter configurations, \mbox{PB-MCTS} and H-MCTS reach similar win rates.}
	\label{fig:results}
\end{figure}

	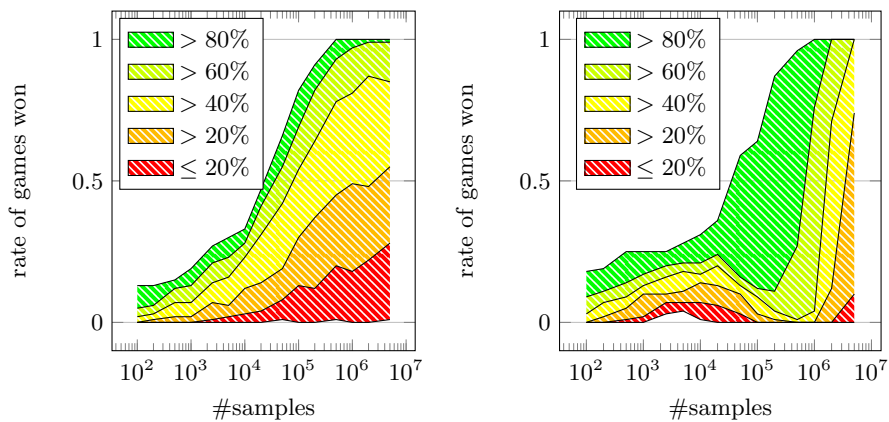
\begin{figure}[th!]
		\subfloat[Configuration percentiles of PB-MCTS]{	
		\begin{tikzpicture}
\begin{semilogxaxis}[
	height = 0.5*\linewidth,
	width = 0.46*\linewidth,
	xlabel={\#samples},
	ylabel={rate of games won},
	legend style = { at = {(0.5,0.98)}},
   ymajorgrids,
	cycle list name=my black white
]

\addplot[name path=1,forget plot] coordinates{
(100,0.13)
(200,0.13)
(500,0.15)
(1000,0.19)
(2500,0.27)
(5000,0.3)
(10000,0.33)
(20000,0.47)
(50000,0.66)
(100000,0.82)
(200000,0.91)
(500000,1)
(1000000,1)
(2000000,1)
(5000000,1)
};
\addplot[name path=3,forget plot] coordinates{
(100,0.05)
(200,0.06)
(500,0.12)
(1000,0.13)
(2500,0.21)
(5000,0.23)
(10000,0.28)
(20000,0.41)
(50000,0.55)
(100000,0.69)
(200000,0.82)
(500000,0.93)
(1000000,0.97)
(2000000,0.99)
(5000000,0.99)

};
\addplot[name path=5,forget plot] coordinates{
(100,0.02)
(200,0.03)
(500,0.07)
(1000,0.07)
(2500,0.14)
(5000,0.16)
(10000,0.23)
(20000,0.31)
(50000,0.42)
(100000,0.54)
(200000,0.64)
(500000,0.78)
(1000000,0.81)
(2000000,0.87)
(5000000,0.85)

};
\addplot[name path=7,forget plot] coordinates{
(100,0)
(200,0.01)
(500,0.02)
(1000,0.02)
(2500,0.07)
(5000,0.06)
(10000,0.12)
(20000,0.14)
(50000,0.19)
(100000,0.3)
(200000,0.37)
(500000,0.45)
(1000000,0.49)
(2000000,0.48)
(5000000,0.55)

};
\addplot[name path=9,forget plot] coordinates{
(100,0)
(200,0)
(500,0)
(1000,0)
(2500,0.01)
(5000,0.02)
(10000,0.03)
(20000,0.04)
(50000,0.08)
(100000,0.13)
(200000,0.12)
(500000,0.2)
(1000000,0.18)
(2000000,0.22)
(5000000,0.28)

};
\addplot[name path=11,forget plot] coordinates{
(100,0)
(200,0)
(500,0)
(1000,0)
(2500,0)
(5000,0)
(10000,0)
(20000,0)
(50000,0.01)
(100000,0)
(200000,0)
(500000,0.01)
(1000000,0)
(2000000,0)
(5000000,0.01)

};

\addplot[pattern=custom north west lines,hatchspread=3pt,hatchthickness=1.5pt,hatchcolor=b9]fill between[of=1 and 3];
\addplot[pattern=custom north west lines,hatchspread=3pt,hatchthickness=1.5pt,hatchcolor=b8]fill between[of=3 and 5];
\addplot[pattern=custom north west lines,hatchspread=3pt,hatchthickness=1.5pt,hatchcolor=b7]fill between[of=5 and 7];
\addplot[pattern=custom north west lines,hatchspread=3pt,hatchthickness=1.5pt,hatchcolor=b5]fill between[of=7 and 9];
\addplot[pattern=custom north west lines,hatchspread=3pt,hatchthickness=1.5pt,hatchcolor=b1]fill between[of=9 and 11];

\legend{$>80\%$,$>60\%$,$>40\%$,$>20\%$,$\leq 20\%$}
\end{semilogxaxis}
\end{tikzpicture}		
		\label{fig:rucbPercentiles}}		
		\subfloat[Configuration percentiles of H-MCTS]{\begin{tikzpicture}
\begin{semilogxaxis}[
	height = 0.5*\linewidth,
	width = 0.48*\linewidth,
	xlabel={\#samples},
	ylabel={rate of games won},
	legend style = { at = {(0.5,0.98)}},
   ymajorgrids,
	cycle list name=my black white
]

\addplot[name path=1,forget plot] coordinates{
(100,0.18)
(200,0.19)
(500,0.25)
(1000,0.25)
(2500,0.25)
(5000,0.28)
(10000,0.31)
(20000,0.36)
(50000,0.59)
(100000,0.64)
(200000,0.87)
(500000,0.96)
(1000000,1)
(2000000,1)
(5000000,1)

};
\addplot[name path=3,forget plot] coordinates{
(100,0.09)
(200,0.11)
(500,0.14)
(1000,0.17)
(2500,0.2)
(5000,0.21)
(10000,0.21)
(20000,0.24)
(50000,0.16)
(100000,0.12)
(200000,0.11)
(500000,0.27)
(1000000,0.76)
(2000000,1)
(5000000,1)

};
\addplot[name path=5,forget plot] coordinates{
(100,0.03)
(200,0.07)
(500,0.09)
(1000,0.13)
(2500,0.16)
(5000,0.18)
(10000,0.17)
(20000,0.2)
(50000,0.13)
(100000,0.09)
(200000,0.04)
(500000,0.01)
(1000000,0.04)
(2000000,0.71)
(5000000,1)

};
\addplot[name path=7,forget plot] coordinates{
(100,0)
(200,0.02)
(500,0.05)
(1000,0.1)
(2500,0.1)
(5000,0.11)
(10000,0.14)
(20000,0.13)
(50000,0.1)
(100000,0.03)
(200000,0.01)
(500000,0)
(1000000,0)
(2000000,0.12)
(5000000,0.74)

};
\addplot[name path=9,forget plot] coordinates{
(100,0)
(200,0)
(500,0.01)
(1000,0.02)
(2500,0.07)
(5000,0.07)
(10000,0.07)
(20000,0.06)
(50000,0.03)
(100000,0)
(200000,0)
(500000,0)
(1000000,0)
(2000000,0)
(5000000,0.1)

};
\addplot[name path=11,forget plot] coordinates{
(100,0)
(200,0)
(500,0)
(1000,0)
(2500,0.03)
(5000,0.04)
(10000,0.01)
(20000,0)
(50000,0)
(100000,0)
(200000,0)
(500000,0)
(1000000,0)
(2000000,0)
(5000000,0)
};

\addplot[pattern=custom north west lines,hatchspread=3pt,hatchthickness=1.5pt,hatchcolor=b9]fill between[of=1 and 3];
\addplot[pattern=custom north west lines,hatchspread=3pt,hatchthickness=1.5pt,hatchcolor=b8]fill between[of=3 and 5];
\addplot[pattern=custom north west lines,hatchspread=3pt,hatchthickness=1.5pt,hatchcolor=b7]fill between[of=5 and 7];
\addplot[pattern=custom north west lines,hatchspread=3pt,hatchthickness=1.5pt,hatchcolor=b5]fill between[of=7 and 9];
\addplot[pattern=custom north west lines,hatchspread=3pt,hatchthickness=1.5pt,hatchcolor=b1]fill between[of=9 and 11];

\legend{$>80\%$,$>60\%$,$>40\%$,$>20\%$,$\leq 20\%$}
\end{semilogxaxis}
\end{tikzpicture}\label{fig:uctPercentiles}}
		
		\caption{The distribution of hyperparameters to wins is shown in steps of $0.2$ percentiles. The amount of wins decreases rapidly for H-MCTS if the parameter setting is not among the best $20\%$. On the other hand, PB-MCTS shows a more robust curve without such a steep decrease in win rate.}
		\label{fig:results2}
	\end{figure}

\subsection{Parameter Settings}

Both algorithms H-MCTS and PB-MCTS are subject to the following hyperparameters:
\begin{itemize}
	\item \emph{Rollout length}: the number of actions performed at most per rollout (tested with: $5$, $10$, $25$, $50$).
	\item \emph{Exploration-exploitation trade-off}: the $C$ parameter for \mbox{H-MCTS} and the $\alpha$ parameter for PB-MCTS (tested with: $0.1$ to $1$ in $10$ steps).
	\item \emph{Allowed transition-function samples per move (\#samples)}: a hardware-independent parameter to limit the time an agent has per move\footnote{
	Please note that this is a fair comparison between PB-MCTS and H-MCTS: The first uses more \#samples per iteration, the latter uses more iterations.} (tested with logarithmic scale from $10^2$ to $5\cdot 10^6$ in $10$ steps).
	
\end{itemize}
For each combination of parameters $100$ runs are executed.
We consider \#samples to be a parameter of the problem domain, as it relates to the available computational resources. 
The rollout length and the trade-off parameter are optimized.

\section{Results}
\label{sec:results}

PB-MCTS seems to work well if tuned, but showing a more steady but slower convergence rate if untuned, which may be due to the exponential growth.

\subsection{Tuned: Maximal Performance}
Fig.~\ref{fig:results} shows the maximal win rate over all possible hyperparameter combinations, given a fixed number of transition-function samples per move.
One can see that for a lower number of samples ($\leq 1000$),
both algorithms lose most games, but H-MCTS has a somewhat better performance in that region. However, 
Above that threshold, H-MCTS no longer outperforms PB-MCTS.
In contrary, PB-MCTS typically achieves a slightly better win rate than H-MCTS.

\subsection{Untuned: More robust but slower}
We also analyzed 
the distribution of wins for non-optimal hyper-parameter configurations. 
Fig.~\ref{fig:results2} shows several curves of win rate over the number of samples, each representing a different percentile of the distribution of the number of wins over the hyperparmenter configurations.
The top lines of Fig.~\ref{fig:results2} correspond to the curves of Fig.~\ref{fig:results}, since they show the results of the the optimal hyperparameter configuration.
Below, we can see how non-optimal parameter settings perform.
For example, the second line from the top shows the $80\%$ percentile, i.e. the configuration for which $20\%$ of the parameter settings performed better and $80\%$ perform worse, calculated independently for each sample size.
For PB-MCTS (top of Fig.~\ref{fig:results2}), the $80\%$ percentile line lies next to the optimal configuration from Fig.~\ref{fig:results}, whereas for H-MCTS there is a considerable gap between the corresponding two curves.
%
%
In particular, the drop in the number of wins around $2 \cdot 10^5$ samples is notable. 
Apparently, H-MCTS gets stuck in local optima for most hyperparameter settings.
PB-MCTS seems to be less susceptible to this problem because its win count does not decrease that rapidly.

On the other hand, untuned PB-MCTS seems to have a slower convergence rate than untuned H-MCTS, as can be seen for high \#sample values. This may be due to the exponential growth of trajectories per iteration in PB-MCTS.



\section{Conclusion}
\label{sec:conclusion}
In this paper, we proposed PB-MCTS, a new variant of Monte Carlo tree search which is able to cope with preference-based feedback. In contrast to conventional MCTS, this algorithm uses relative UCB as its core component. We showed how to use trajectory preferences in a tree search setting by performing multiple rollouts and comparisons per iteration. 

Our evaluations in the 8-puzzle domain showed that the performance of H-MCTS and PB-MCTS strongly depends on adequate hyperparameter tuning.
PB-MCTS is better able to cope with suboptimal parameter configurations and erroneous heuristics for lower sample sizes, whereas H-MCTS has a better convergence rate for higher values. 


One main problem with preference-based tree search is the exponential growth in the number of explored trajectories.
Using RUCB grants the possibility to exploit only if both actions to play are the same.
This way the exponential growth can be reduced.
But nevertheless we are currently working on techniques that allow to prune the binary subtree 
without
changing the feedback obtained in each node. Motivated by alpha-beta pruning and similar techniques in conventional game-tree search, we expect that such techniques can further improve the performance 
and remove the exponential growth to some degree.

\subsubsection{Acknowledgments.}
This work was supported by the German Research Foundation (DFG project number FU 580/10). 
We gratefully acknowledge the use of the Lichtenberg high performance computer of the TU Darmstadt for our experiments.



\bibliographystyle{splncs04}
\end{document}